\documentclass[sigconf,authorversion,nonacm]{acmart}
\AtBeginDocument{%
  }

\newtoggle{twocolumnformat}

\toggletrue{twocolumnformat}

\acmConference[IUI '23]{Proceedings}{March 27-31,
  2023}{Sydney, Australia}
\acmPrice{15.00}
\acmISBN{978-1-4503-XXXX-X/18/06}

\usepackage{mathtools} 
\usepackage{todonotes}
\usepackage{cuted} 
	\newcommand*\rfrac[2]{{}^{#1}\!/_{#2}}
	
\copyrightyear{2023} 
\acmYear{2023} 
\setcopyright{rightsretained} 
\acmConference[IUI '23]{28th International Conference on Intelligent User Interfaces}{March 27--31, 2023}{Sydney, NSW, Australia}
\acmBooktitle{28th International Conference on Intelligent User Interfaces (IUI '23), March 27--31, 2023, Sydney, NSW, Australia}
\acmDOI{10.1145/3581641.3584050}
\acmISBN{979-8-4007-0106-1/23/03}

\begin{document}

\title{Improving Fairness in Adaptive Social Exergames via Shapley Bandits}
\author{Robert C. Gray}
\affiliation{%
	\institution{Department of Digital Media \\ Drexel University}
	\city{Philadelphia, PA}
	\country{USA}}
\email{robert.c.gray@drexel.edu}

\author{Jennifer Villareale}
\affiliation{%
	\institution{Department of Digital Media \\ Drexel University}
	\city{Philadelphia, PA}
	\country{USA}}
\email{jmv85@drexel.edu}

\author{Thomas B. Fox}
\affiliation{%
	\institution{Department of Digital Media \\ Drexel University}
	\city{Philadelphia, PA}
	\country{USA}}
\email{seafordtbf@gmail.com}

\author{Diane H. Dallal}
\affiliation{%
    \institution{Department of Psychiatry\\ University of Pennsylvania}
	\city{Philadelphia, PA}
	\country{USA}}
\email{dianehdallal@gmail.com}

\author{Santiago Ontañón}
\affiliation{%
	\institution{Google Research}
	\city{Philadelphia, PA}
	\country{USA}}
\email{santi.ontanon@gmail.com}

\author{Danielle Arigo}
\affiliation{%
	\institution{Department of Psychology \\ Rowan University}
	\city{Glassboro, NJ}
	\country{USA}}
\email{arigo@rowan.edu}

\author{Shahin Jabbari}
\affiliation{%
	\institution{Department of Computer Science \\ Drexel University}
	\city{Philadelphia, PA}
	\country{USA}}
\email{shahin@drexel.edu}

\author{Jichen Zhu}
\affiliation{%
  \institution{Department of Digital Design \\ IT University of Copenhagen}
  \city{Copenhagen}
  \country{Denmark}}
\email{jichen.zhu@gmail.com}

\renewcommand{\shortauthors}{Gray et al.}

\begin{abstract}

Algorithmic fairness is an essential requirement as AI becomes integrated in society. In the case of social applications where AI distributes resources, algorithms often must make decisions that will benefit a subset of users, sometimes repeatedly or exclusively, while attempting to maximize specific outcomes. How should we design such systems to serve users more fairly? This paper explores this question in the case where a group of users works toward a shared goal in a social exergame called {\em Step Heroes}. We identify adverse outcomes in traditional multi-armed bandits (MABs) and formalize the {\em Greedy Bandit Problem}. We then propose a solution based on a new type of fairness-aware multi-armed bandit, {\em Shapley Bandits}. It uses the Shapley Value for increasing overall player participation and intervention adherence rather than the maximization of total group output, which is traditionally achieved by favoring only high-performing participants. We evaluate our approach via a user study (\textit{n}=46). Our results indicate that our Shapley Bandits effectively mediates the Greedy Bandit Problem and achieves better user retention and motivation across the participants.
\end{abstract}



\maketitle

\section{Introduction}

Many organizations are deploying AI systems to manage and allocate resources, such as assigning work tasks and distributing work schedules to personnel to improve productivity in dynamic environments~\cite{naveh2007workforce, jaumard1998generalized, millar1998cyclic}. However, these opportunities have also raised new challenges, most notably the potential for AI systems to unfairly allocate opportunities and resources among a group of users. In the case of social applications where users work together toward a shared goal, this challenge is particularly difficult; individual user contributions to the group may vary, and the AI may place a subset of users above others to maximize outcomes without considering user contributions. 
%
%
%
Hence, fairness has become an emerging and important research area within the AI and machine learning communities~\cite{BerkHJKR18, fairmlbook}. Work in this area identified fairness problems in a wide range of AI techniques such as classification~\cite{HardtPS16, ZafarVGG2017}, regression~\cite{BerkHJJKMNR17}, online learning~\cite{JosephKMR16, LiuRDMP17}, or reinforcement learning~\cite{JabbariJKMR17}. 
Despite a growing body of work, we do not yet have good solutions for fairness-aware AI systems that need to distribute resources in social applications where groups of users work together towards a joint goal~\cite{zhu2020player,zhu2019experience}, such as social exergames.


In this paper, based on prior work in an AI-based adaptive social fitness application~\cite{rabbi2015mybehavior, yom2017encouraging, zhou2018evaluating}, we found that commonly used metrics to capture individual users' contributions to the group were insufficient due to fairness issues. 
We noticed an AI tasked with adapting the experience to maximize users' physical activity (PA) repeatedly placed some users' preferences above others, leading to non-adherence by marginalized users. 
Rather than focusing purely on the amount of PA (e.g., daily steps) achieved by the group, this paper examines the potential benefits of a fairness-aware approach targeting player adherence to the group activity as an alternative metric for the AI to maximize. Specifically, we present a new type of fairness-aware multi-armed bandit approach that we call the {\em Shapley Bandit}, as it uses the Shapley Value~\cite{shapley1953} to enforce fairness constraints. With this approach, we explore how to design AI for social applications that can serve a group of users while ensuring fundamental expectations of ``distributive fairness'' are maintained~\cite{cohen1987distributive, alexander1987role}. 
Compared to prior work in fairness in online learning and, in particular, multi-armed bandit problems~\cite{JosephKMR16, LiuRDMP17, LiLJ19}, which assume that the choices made by the AI only affect one of the individuals, our work studies the setting where the choices made by the AI can affect more than one individual.

We conducted a mix-methods user study to evaluate the effectiveness and tradeoffs of our Shapley Bandits compared to traditional bandits in the social exergame \textit{Step Heroes}. Our user study results (\textit{n}=46) show that our Shapley Bandits achieve higher motivation scores for physical activity than a control group, whereas a traditional bandit approach 
did not. Additionally, it is better at retaining users, where users are more likely to maintain adherence to the intervention. 
Based on our findings, we propose implications for \textit{behavioral fairness} in AI-driven social applications in terms of considering users' contributions to the group as a factor for fairness. 

Our primary contribution is a notion of fairness integrated into an AI strategy that makes its selections based on the disparity between a user's exerted effort within a team endeavor and the degree of favorable treatment they receive from the AI in a multiplayer environment. 
Specifically, this work brings the following contributions: 
\begin{itemize}
    \item To the best of our knowledge, this is the first attempt in the game AI research community to model group preferences and the related fairness issues.
    \item We present a new fairness-aware multi-armed bandit strategy, the {\em Shapley Bandit}, that uses the Shapley Value to estimate a notion of {\em disparity} and then aims to reduce it while at the same time maximizing expected reward.
    \item We present empirical evidence of the effectiveness of our approach in the context of exergames via a user study (\textit{n}=46).
\end{itemize}

Our research indicates the importance of integrating fairness concerns in multi-armed bandits literature, especially in the context of adaptive social exergames that serve multiple players. Our empirical results show that if the AI only focuses on total group output (e.g., a group's total daily steps), it risks consistently favoring only high-performing participants while ignoring the needs of marginalized users. Over time, this traditional approach may lead to the latter dropping out, thus raising concerns about the ethics as well as the effectiveness of the intervention. We observed that even though introducing fairness considerations does not directly lead to higher PA levels, it contributes to user retention and adherence, which is critical to lasting behavior change. Therefore, holistic and ethical considerations of the AI's utility function are key design issues in such applications and need further research.

\section{Related Work}


\subsection{Fairness in AI}
With the rapid adaptation of AI systems in domains such as financial lending, sentencing, and healthcare, there have been growing concerns about societal aspects such as bias and transparency when deploying these systems. These concerns have led to an emerging research area within the AI and machine learning communities known as algorithmic fairness (see~\cite{BerkHJKR18, fairmlbook} for recent surveys). The research in this field typically considers well-studied learning settings such as classification~\cite{HardtPS16, ZafarVGG2017}, regression~\cite{BerkHJJKMNR17}, online learning~\cite{JosephKMR16, LiuRDMP17}, or reinforcement learning~\cite{JabbariJKMR17}. This research examines how bias or discrimination can happen in these settings, proposes formal definitions for what bias or unfairness means mathematically, and designs provably fair algorithmic frameworks. The main focus of the field has been on group fairness, where individuals are divided into groups using a combination of sensitive attributes such as race or gender and where fairness guarantees are designed to apply at the group level~\cite{HardtPS16, BerkHJJKMNR17, Corbett-DaviesP17, ZafarVGG2017}. However, group fairness does not provide any meaningful guarantee at the individual level, still allowing unfair treatment of individuals. Several works have studied individual fairness~\cite{DworkHP+12, JosephKMR16}, and it is known that individual fairness is often hard to enforce in practice~\cite{DworkHP+12}.

Additionally, past research has shown that there is a disconnect between the needs and realities of high-stake applications (like algorithmically informed public decision-making in taxation, justice, or child protection) and tools and research in algorithmic fairness~\cite{veale2018fairness}. This remains true, despite recent efforts in this direction. For example, Richardson et al. studied the gap between the needs of practitioners and the tools offered by fairness research~\cite{richardson2021towards}, and Nakao et al. studied how to directly engage end-users in identifying and addressing fairness issues in AI applications~\cite{nakao2022towards}.

Our work studies individual fairness in online learning and, in particular, multi-armed bandit problems. Prior work~\cite{JosephKMR16, LiuRDMP17, LiLJ19} assumes that the choice of action in each round only affects one of the individuals favorably. Our work differs from prior work in that the choice of action can affect more than one individual either favorably or unfavorably. 

\subsection{Group Fairness in Resource Allocation}\label{sec:rw-group-fairness}

In applications where an AI adapts an experience for a group of human users, a body of work has been developed on using AI techniques to improve productivity through managing and allocating resources~\cite{skobelev2018towards, aubin1992scheduling, spyropoulos2000ai, chow1993knowledge, gierl1993knowledge, robert2020designing}. 
In particular, an important research area is developing AI systems to mediate resources among a group of people, such as assigning workers to tasks~\cite{naveh2007workforce, atabakhsh1991survey}, allocating work schedules~\cite{dowsland1998nurse, jaumard1998generalized, millar1998cyclic}, and matching project requirements with employees' skill sets~\cite{naveh2007workforce}. These decisions are usually affected by a set of \textit{externally} defined objectives and constraints, such as meeting deadlines and maximizing resource utilization~\cite{atabakhsh1991survey}, hours an operator can work~\cite{chow1993knowledge}, and human factors, such as expertise and skill level~\cite{naveh2007workforce}.

Recently, researchers have discussed fairness-related harms that can occur, such as people’s individual experiences with AI systems~\cite{robert2020designing} or how AI systems represent individuals in groups~\cite{ajunwa2019platforms}. Of note, fairness in group AI (i.e., AI managing groups of people by distributing resources) is a particularly difficult challenge, as AI systems can unfairly allocate opportunities, resources, or information~\cite{bird2020fairlearn}. For example, an automated scheduling system may inadvertently withhold jobs from a particular individual due to how they are labeled or represented in the system.

Researchers have focused on fairness by mitigating disparities in the treatment of individuals based on categories such as racial, gender, or age groups~\cite{friedler2019comparative, heidari2019moral, chen2019fairness}. In the case of this research, in which we provide a game intervention toward motivating people to walk more in their daily lives, targeting fairness based on such categories lies outside the scope of our domain and the visibility of our AI. Instead, we aim to address the fundamental issue of ``distribution fairness''~\cite{cohen1987distributive, alexander1987role}, which considers users' utility to the group as a factor for fairness. We address distribution fairness through the lens of Organizational Justice Theory~\cite{greenberg1990organizational, greenberg2013handbook, robert2020designing} extending from J.S. Adams' equity theory~\cite{adams1963towards, adams1965inequity}. The primary assertion of these theories is that there exists an implied contract that individuals participating in a group should receive a distribution of reward commensurate with their personal investment toward the goals of that group, with an emphasis on the negative responses of individuals that are likely to result from a failure to uphold this tenet.


\subsection{Fairness in Group Recommender Systems} \label{sect:recSys}

A closely related area to our research is group recommender systems. A group recommender system makes a decision, often by suggesting items to a group of people engaged in a {\em group activity}~\cite{kim2010group}, such as watching a movie together. They have been used to recommend music~\cite{mccarthy1998musicfx, chao2005adaptive, crossen2002flytrap}, movies and TV~\cite{o2001polylens, yu2006tv}, and restaurant and travel~\cite{mccarthy2002pocket, salamo2012generating} to a group of users collectively. In order to capture the preference of the group as a whole, group recommender systems use group aggregation models to integrate the preferences of individual members~\cite{Dara2020}. 
Fairness is an essential consideration of group recommender systems because not all recommended items satisfy the group members equally. In other words, a group aggregation model decides whose preferences or needs should be given more weight at a given time. 

Group aggregation models can be divided into three main strategies: 1) majority-based, 2) consensus-based, and 3) borderline strategies~\cite{kavsvsak2016personalized, senot2010analysis, masthoff2004group}. Generally speaking, majority-based strategies select the item preferred by most group members. Consensus-based strategies aggregate the preferences of all members somehow and then select items based on the common opinion of the group. Widely used models in this type include the average strategy, least misery strategy, Borda count, most pleasure strategy, average without misery strategy, and fairness strategy. Borderline strategies are in-between the previous two types. An example is the ``dictatorship'' strategy, where items are selected based on one selected member's preferences. 

Most of the above approaches assume that user preferences do not interact with each other. For example, at a party where people listen to the same music, what a person would like to hear next can be influenced by the songs recommended earlier. In this paper, we explore a different way of aggregating user preferences or needs by minimizing the disparity between how much an individual contributes to the group (estimated via the Shapley Value) and the degree to which the AI caters to that player with its recommendations.

\subsection{Personalized Digital Interventions for PA and Social Comparison}\label{sec:background-exergames}
There is a broad base of existing literature on personalized interventions to promote physical activity (PA), especially to adapt PA goals for improving goal adherence~\cite{parkka2010personalization, bleser2015personalized, bull1999effects, rabbi2015automated, arigo2020social}. Among existing work on PA personalization, an emergent approach is to use AI techniques to personalize digital interventions automatically~\cite{paredes2014poptherapy, yom2017encouraging, forman2019can}. A benefit of AI-based interventions is the ability to provide real-time tailoring of feedback and adaptation of goals based on continuous data monitoring on an individual basis. 

Most notably, personalized digital interventions have shown to be more beneficial for behavior change over generic automated systems~\cite{rabbi2015mybehavior}, such as helping individuals develop attainable step goals through more personalized recommendations~\cite{rabbi2015mybehavior, yom2017encouraging, zhou2018evaluating}. However, relatively little work has leveraged AI systems to enhance PA for groups as a whole. Instead, personalization has prioritized the content and timing of individual activity recommendations using demographic information~\cite{buttussi2008mopet}, behavioral patterns~\cite{lin2012bewell}, time and location~\cite{stahl2008mobile, buttussi2008mopet}, and personality~\cite{arteaga2010mobile}.

\begin{figure*}[t!]
	\centering
    \includegraphics[width=0.45\textwidth]{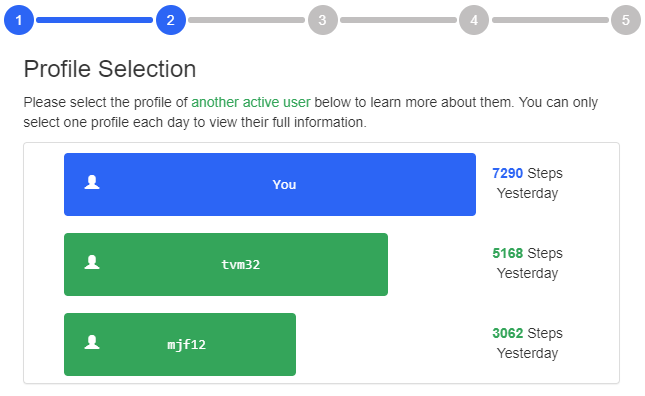}
	\includegraphics[width=0.45\textwidth]{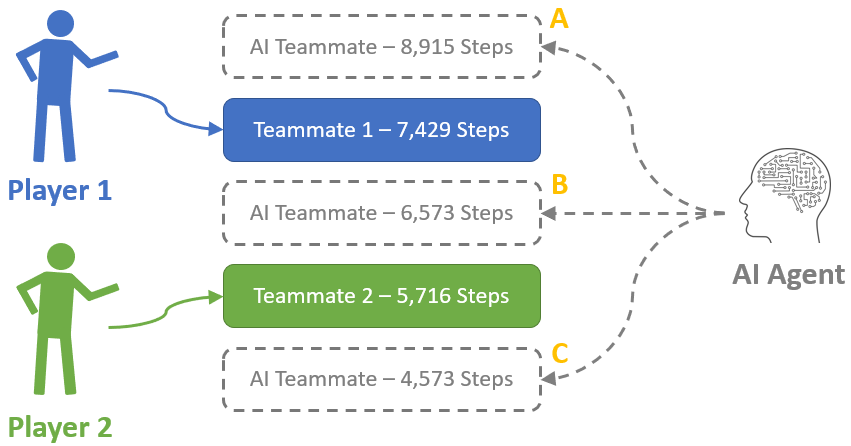}
	\caption{Left: Screenshot of the study profile selection screen in our preliminary study. Right: Illustration of the decision the AI agent makes. The MAB-based AI determines the placement of the third teammate's steps among three options (A, B, or C) to cater to the users' social comparison orientation (SCO) and increase user motivation.}
	\label{fig:pre_study_2_2}
\end{figure*}

Social comparison describes the process by which individuals evaluate themselves or their behavior relative to others~\cite{festinger1954}. In digital group interventions, applications typically use social features such as ranked leaderboards and competitive challenges to engage {\em social comparison} processes~\cite{wood1989}. While these interventions have motivated health behavior change in weight control and PA promotion~\cite{olander2013most, leahey2010effect}, not all users may be equally motivated by these features. Research has shown that individuals prefer different people to compare with~\cite{bennenbroek2002social}, and these preferences may vary over short periods~\cite{arigo2020social}. Comparisons to others who seem ``better off'' than the comparer in a given domain are {\em upward comparisons}, and comparisons to others who seem ``worse off'' are {\em downward comparisons}~\cite{buunk1995comparison}. Unfortunately, most personalized group interventions do not account for individual differences and provide all users with the same kind of generic exposure to social comparison targets~\cite{arigo2020social}, which can negate any need for behavior change efforts~\cite{arigo2015addressing, merchant2017face, wills1981downward}.

An alternative approach is to motivate PA through social gamified fitness applications or ``social exergames'' using similar group features described above. This active research area integrates physical exertion with digital games~\cite{mandryk2016exertion} to transform physical exercise in an engaging game environment and leverage social interactions through facilitating cooperative and competitive game settings to increase and sustain players' motivation for PA~\cite{caro2018understanding, lin2006fish, chen2014healthytogether}. Social interaction through the lens of cooperation and competition has been a clear motivator for PA activities in these games and has been shown to encourage comparisons between players~\cite{ahtinen2009designing, lin2006fish, chen2014healthytogether}. However, social features such as ranked leaderboards and competitive challenges~\cite{arigo2020social, chen2014healthytogether} have been shown to also risk demotivating people due to how they \textit{compare} themselves to others.  
This paper extends the literature on personalization in digital interventions for PA, especially social exergames, by developing an AI approach that automatically models the social comparison preferences of the group and provides appropriate social comparison opportunities.

\section{Research Platform}

The following introduces our preliminary research platform (Section \ref{sec:pre-test}) that helped us identify the fairness issues that arise when optimizing for group performance (Section \ref{sec:greedy-bandit-problem}). We then introduce the research platform for our primary user study (Section \ref{stepheroes_description}) and the AI approach we designed to address fairness (Section \ref{sec:shapley-bandit}).

\subsection{Preliminary Design: Web-Based App with Standard MAB}\label{sec:pre-test}

Since research has established that social elements are a crucial motivating factor for PA~\cite{zhu2018cscw, campbell2008game, feltz2014, max2016, lin2006fish}, our preliminary work aims to promote PA using adaptive personalized AI inspired by {\em social comparison} theory (see Section~\ref{sec:background-exergames}). We designed a web-based app in which a group of users can compare their daily steps with other users' PA-related profiles (e.g., daily steps, favorite types of exercises, and other PA-related interests)~\cite{zhu2021cscw}. Users' steps were captured by Fitbit and synced automatically with our platform. Every day, the user selects one out of two other users that the MAB recommends to them as (social) comparison targets (Figure~\ref{fig:pre_study_2_2} left). Each group consists of two randomly paired human users and one artificial user whose steps are controlled by the AI, similar to our previous work~\cite{gray2021cog, zhu2021cscw}. The artificial user gives the MAB an additional chance to provide good comparison targets, where human users' daily steps are unpredictable and thus may not provide the motivating comparison targets the two humans in the group need.
Based on the changes in the users' daily steps and their reported motivation for PA, the MAB updates its model for their real-time preference of social comparison direction and reaction to the comparison. It then recommends comparison targets to maximize the chance of increasing PA and motivation.

We conducted a 21-day user study (\textit{n}=53) to investigate our personalization mechanism and how exposure to personalized social comparison targets may affect users' PA and motivation to exercise. Our initial results indicated that our approach was able to automatically model and manipulate social comparison in the pursuit of PA promotion. The detected effects achieved small-to-moderate effect sizes, illustrating the real-world implications of the intervention for enhancing motivation and PA~\cite{zhu2021cscw}.

\begin{figure*}[t!]
	\includegraphics[width=.65\textwidth]{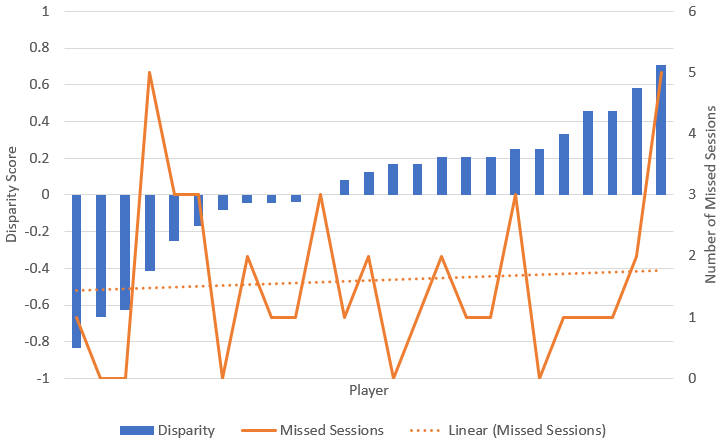}
	\centering
	\caption{Sorted disparity scores (blue) and miss likelihood (orange, secondary axis) for users in the preliminary design study. Users sorted to the left depict those who were given very good treatment relative to their effort (i.e., step performance), while users sorted to the right received poor treatment relative to their efforts. The graph reveals a positive correlation between a user's disparity score and their number of missed sessions (Pearson's $R{=}0.16$).}
	\Description{Disparity scores for users in our preliminary design study}
	\label{fig:shapley-analysis-disparity}
\end{figure*}

\subsection{Problem Formulation: The Greedy Bandit Problem}\label{sec:greedy-bandit-problem}
Further analysis of the results in our preliminary design raised questions regarding how the MAB distributes its ``attention'' to different users. The MAB selects an arm (in the form of an artificial user as the comparison target for two human users) to maximize the sum of steps and self-reported motivation of both users. We note that, in essence, our MAB operates similarly to the average strategy for group recommender systems (Section \ref{sect:recSys}). 

Let us consider Figure~\ref{fig:pre_study_2_2} (right) and imagine a case in which the model predicts Arm A will yield a highly positive effect on User 1 due to their preference for upward comparisons. However, in this example, the AI also predicts a negative effect on User 2 due to User 2's dislike for upward comparisons. It may be that the {\em net} prediction of Arm A still yields the highest overall effect (i.e., the strength of User 1's preference may be larger than that of User 2), so our traditional bandit strategy optimizing for the sum of steps of both users will select it. When faced with the same decision again later, the MAB is likely to make the same decision. Over repeated selections, a blind favoring of User 1 in a ``greedy'' pursuit of a specific metric may lead to overlooking User 2 regularly. 

In a larger group, a similar case may arise when a particular arm appeals to the majority of users (e.g., 60\% of users prefer one game setting that the other 40\% dislike). This is similar to the majority-based group aggregation models in group recommender systems (Section~\ref{sect:recSys}). Over time, a greedy strategy may risk repeatedly neglecting the minority, leading to dissatisfaction among the users. 

{\bf {\em Disparity}.} Further analysis of our user study data indeed shows evidence that users whose social comparison preferences are not met by the MAB have a higher chance of missing sessions. In Figure~\ref{fig:shapley-analysis-disparity}, the blue bars show the {\em disparity} between each user's effort and treatment, defined as follows. 
We define the {\em effort} $E_i$ of a user $i$ as their average daily total steps. 
We define a the {\em treatment} $T_i$ of a user $i$ as the total number of days where they received the best social comparison targets according to the MAB estimations. We now define 
$PR(E_i)$ as the percentile where user $i$ falls in terms of effort across all users in the study (e.g., 1.0 if they are the one with the most steps, and 0.0 if they are the one with the least steps) and $PR(T_i)$ as the percentile where the user $i$ falls in terms of treatment. The disparity of a user $i$ is then defined as $D_i=(PR(E_i)-PR(T_i))$. 
This calculation results in a number between potential extremes of -1 (i.e., highest treatment with lowest effort) and +1 (i.e., lowest treatment with highest effort). 
Figure~\ref{fig:shapley-analysis-disparity} sorts the users from least disparity to the most. The orange plot shows the corresponding user's total number of missed sessions (days). 

Our results show a positive correlation (Pearson's $R{=}0.16$), where users who receive lower treatment relative to their effort percentile are more likely to miss sessions. In other words, high {\em disparity} is correlated with non-adherence behavior.
When the AI does not make choices catering to a user's preferences to a degree commensurate to that user's investment of effort within the game to support their team, we see that those users are more likely to miss sessions. In other words, they are more likely to divest from the activity and the intervention it provides. 

We argue that this result reveals two problems. The first is that when a user ceases to participate in the application in which the AI agent operates, any intervention attempted by the AI agent becomes completely ineffective.
The second involves a limited notion of ``fairness'' derived from the example above. Although a user or group of users may be in the preferential minority, they still bring value to the game environment and therefore contribute to the overall game experience and community. Indeed, even a user's mere presence or participation enables the multiuser environment to exist for others.
Therefore, we would like to promote a best practice in which an AI agent operating in a group setting recognizes this contribution provided by each user and ensures it is rewarded.

We refer to this problem as the {\em Greedy Bandit Problem}. Our main technical contribution, the {\em Shapley Bandit} (Section \ref{sec:shapley-bandit}), is designed to directly target this problem via the following approach: (1) using the Shapley Value to get an estimate of disparity (notice the definition for disparity above relies on having access to the data for the whole user study, and hence it cannot be directly used within in the bandit), and (2) using a fairness-aware bandit strategy targeted at reducing disparity.

\subsection{Improved Design: Idle Social Exergame {\em Step Heroes}}
\label{stepheroes_description}


\begin{figure*}[t!]
    \includegraphics[width=0.8\textwidth]{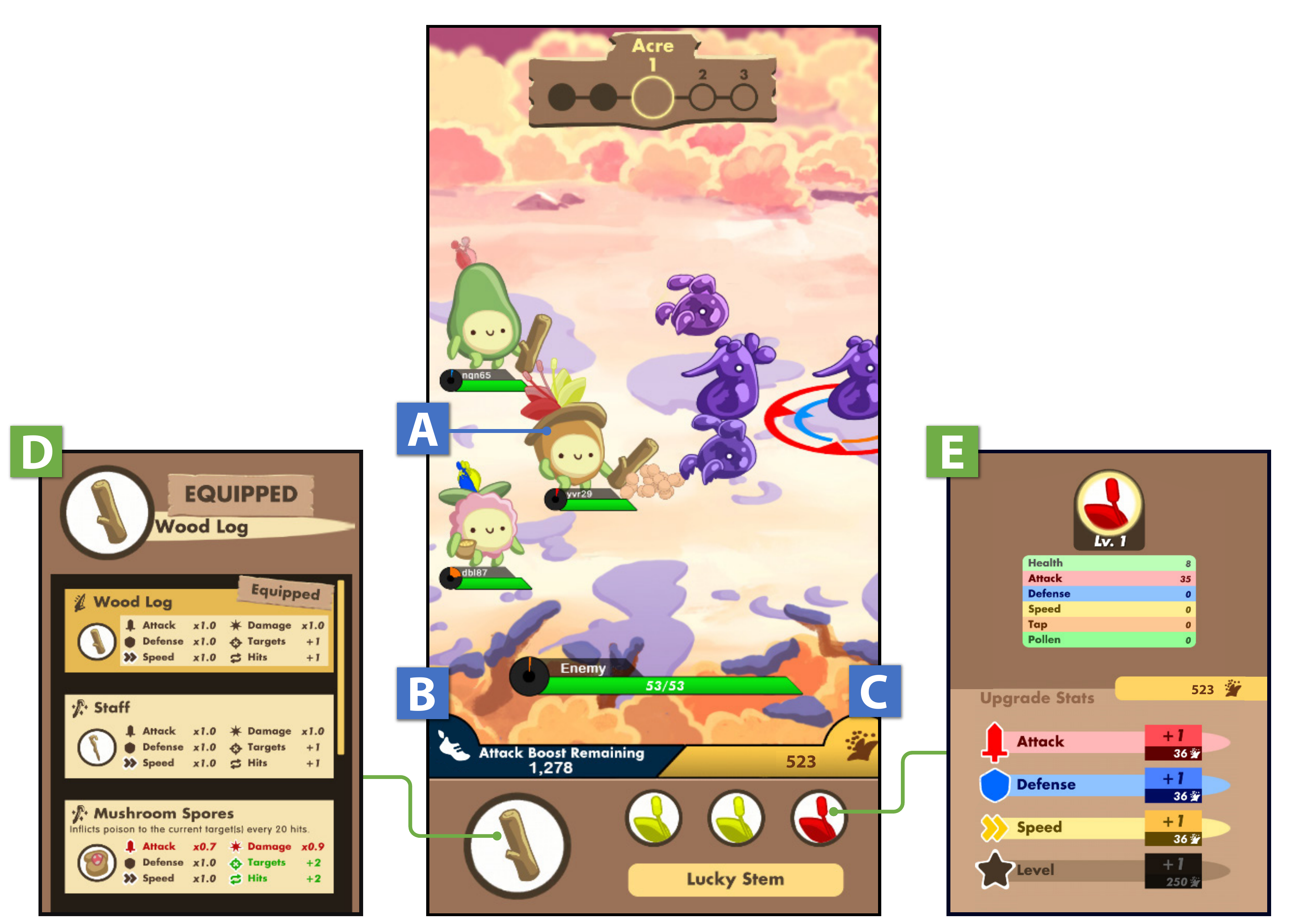}
	\centering
	\caption{Overview of the user interface of {\em Step Heroes}. Part A illustrates the unique avatars representing each player in the game environment. Part B showcases the attack boost meter in which players can contribute ``steps'' to boost their avatar's attack power. Part C shows the currency meter, consisting of ``pollen'' earned from defeating enemies and spent on avatar upgrades to their specific stats (Part E). Players may also purchase and equip a variety of available weapon options that each have unique effects and abilities for further customization of their avatar, shown in part D.}
	\Description{User interface overview for Step Heroes}
	\label{fig:stepheroes_UI}
\end{figure*}

For the main user study presented in this paper, we implemented the same mechanism of adapting the social comparison environment of two users by MAB, this time within the context of a mobile game called \textit{Step Heroes}. There is a large body of work that shows incorporating game-like elements is effective in motivating physical activity~\cite{lin2006fish, chen2014healthytogether, campbell2008game}. In particular, \textit{Step Heroes} was designed to include idle game features, which can be especially effective in behavior change games~\cite{villareale2019idle, alharthi2018, juul2010}.

{\em Step Heroes} is a multiplayer exergame where teams of three players work together to combat invasive enemies who are draining the health of the forest. The goal of the game is, as a team, to clear as many forest areas as possible by defeating the enemies living there. 
Similar to many idle games, a main part of the gameplay in {\em Step Heroes} centers around upgrading avatars. A player needs to decide how to allocate her currency (based on her real-world steps and the experience points she earned from defeating past enemies) to develop the different combat abilities of her avatar. To be really successful, she also needs to work together with her teammates so that their avatars have complementary strengths in the joint battles (Figure~\ref{fig:stepheroes_UI}, middle panel). 

A critical design decision was to bring the avatar upgrade mechanics to the center of players' collaboration and social comparison. Each avatar has a blossom over its head (Figure~\ref{fig:stepheroes_UI}, part A). The blossoms visually expand and contract according to the number of steps each player has taken, providing a visual representation of each player's contribution to the team for easy social comparison. In addition, the color of an avatar's blossom reflects its combat abilities. For example, if a player invests heavily in her avatar's speed and attack abilities, its blossom will show yellow and red. 
This design choice was intended to summarize a player's gameplay decisions and help other players to decide how to upgrade their avatars for a more balanced team composition.

{\em Step Heroes} is designed for users to experience in regular short bursts of play every day over several weeks. As an idle game, {\em Step Heroes} is designed to encourage players to frequently return to examine their team's progress and to spend their step-based currency on upgrading their avatars~\cite{villareale2019idle}. In addition, we designed the game to support a wide range of engagement to fit within users' daily schedules. A player can accumulate all the steps and play a short session a day or log into the game and play as many times as they like.  



In the rest of the section, we focus on how a player's interaction with the profiles in {\em Step Heroes}---the main intervention that uses social comparison---motivates players' PA. It consists of a three-step process. 
Notice that in order to test our AI, each team consists of two human players and one AI-controlled artificial player. Human players are again not made aware that one of their teammates is an AI player.



\begin{figure}[t!]
	\includegraphics[width=0.49\textwidth]{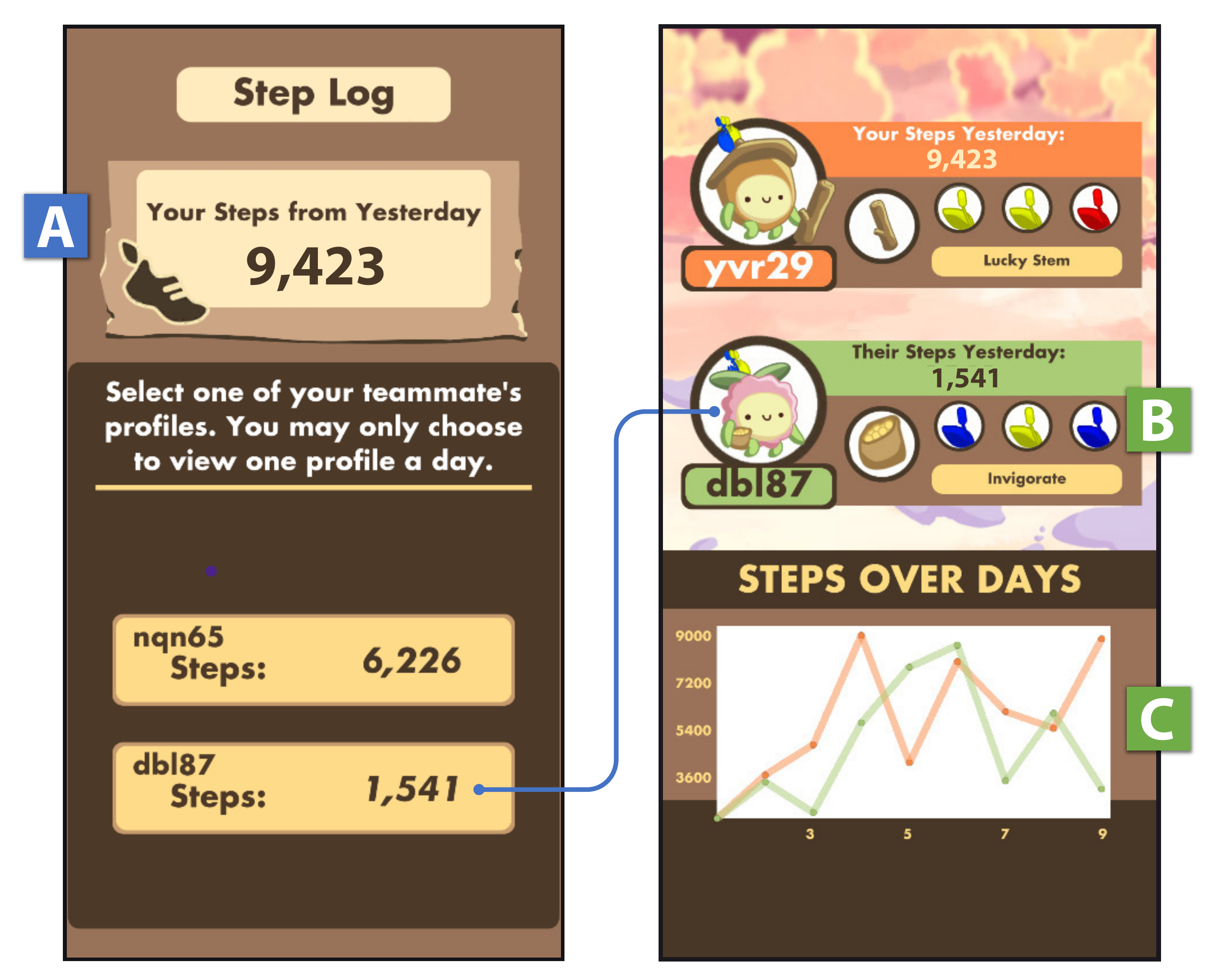}
	\centering
	\caption{The {\em Step Heroes} Teammate Profile Selection screen (left) and the Full Game Profile screen (right) that is revealed when a teammate is chosen.
	}
	\Description{Step Heroes teammate profile selection page}
	\label{fig:game_choicepoint_1_2}
\end{figure}

{\bf {\em Teammate Profile Selection.}} Each day when a player logs in, the game first displays the player's daily step count from the previous day at the top of the interface (automatically retrieved from the player's Fitbit account) and the profiles of both of their teammates (Figure~\ref{fig:game_choicepoint_1_2}, part A). We purposefully designed this selection page with minimal information so the player could focus on a single dimension for social comparison: the previous day's total steps. The player is asked to select between their teammates to continue, with a notification that they can only choose one profile to review {\em in full} each day and cannot go back to choose another.

{\bf {\em Full Player Profile.}} After the player confirms their selection, they are given more detailed information about the selected teammate. The profile reveals further information about that player's game decisions, such as stamen upgrades, weapons, and abilities. For example, players may click on their teammate's weapon or stamen icons (Figure~\ref{fig:game_choicepoint_1_2}, part B) to reveal details about their teammate's actions. Below the game information, we provide users with a step graph to examine and compare their step counts with the selected teammate over time (Figure~\ref{fig:game_choicepoint_1_2}, part C). This screen is the only time users can examine their teammate's profiles. During gameplay, we did not provide access to their teammate's profiles to ensure a meaningful selection at the start of each day.  

{\bf {\em Main Gameplay.}} After the user has explored the selected profile, they may continue to the game interface (center image in Figure~\ref{fig:stepheroes_UI}, where players may begin to play \textit{Step Heroes}). Players can return to the game as often as they would like throughout the day; however, the Teammate Profile Selection screen will only display once during their first login of each day. 

\section{Shapley Bandits: Using Shapley-based Fairness Constraints to Guide Bandits}\label{sec:shapley-bandit}

To address the Greedy Bandit Problem (Section~\ref{sec:greedy-bandit-problem}), we propose a new type of multi-armed bandit~\cite{robbins1952some, thompson1933likelihood} with two new contributions. The first is a method for calculating disparity by making use of the Shapley Value equation (Equation~\ref{eq:shapley}) to estimate the contributions of each player (notice that the definition for disparity in Section~\ref{sec:greedy-bandit-problem} requires having access to the data for the whole user study, and hence can only be used for post-user study analysis). The second is a new type of fairness-aware bandit strategy designed to reduce disparity while at the same time maximizing reward. These two contributions result in what we call a {\em Shapley Bandit}. Let us first introduce the concepts of the Shapley Value and multi-armed bandits before describing Shapley Bandits.

\subsection{Shapley Value}
\label{sec:related-shapley}

The core of our solution for allocating AI attention toward group members is rooted in economic theory, where we leverage the concept of the Shapley Value to determine fair attribution of a group's success to the individuals within that group. The Shapley Value~\cite{shapley1953} was proposed as a method for determining the costs and rewards that ought to be attributed to individuals working together in a group endeavor~\cite{winter2002}. Though the individual performance within a group may be known, the output of the individual members as a result of joining together as a team is more difficult to assess (due to synergies and cross-influences among team members). 
For example, consider a group of people joining together to be physically active (measured in daily walking steps), where each individual walks an average of 10,000 steps daily. Then, a new team member joins who walks 12,000 steps daily, after which the average steps of the other people increase to 11,000 steps per person. The new member's contribution to the team is not just their 12,000 steps but also part of the extra 1,000-step improvement gained by the others (which would not have happened otherwise).

The Shapley Value aids in assessing the total contribution of an individual member toward the group's resulting aggregate performance. Shapley proposed that such a calculation would satisfy the following four axioms (and his solution has been shown to be the only one that will do so~\cite{roth1988, ma2020}): 1) Symmetry, where two individuals are interpreted as equivalent if they prove to be completely interchangeable within the group; 2) Nullity, where an individual who provides no value to any part of the group beyond their individual achievement is given zero synergistic attribution; 3) Additivity, where if separating a game into smaller games (i.e., ``rounds''), the contribution scores from the rounds add up to the score for the whole game; and 4) Efficiency, where the contributions calculated for the individuals add up to the full contribution of the group.
%
%

For a player $i$ participating in a coalition of players $N$, we consider each hypothetical sub-coalition ($S$) of players of $N$ that do not yet contain player $i$ (i.e., $N\setminus\{i\}$). We then add $i$ to each of these sub-coalitions, noting the marginal benefit that player $i$ adds to the group, and average these values. The result is the expected value for the contribution that $i$ brings to $N$ as a whole. This marginal value contributed by $i$, denoted as $\varphi_i(N,v)$, is summarized in the following equation~\cite{ma2020}:  

\begin{equation}\label{eq:shapley}
	\varphi_i(N,v) = \frac{1}{N!}\sum_{S\subseteq N\setminus\{i\}}{|S|!(|N|-|S|-1)![v(S\cup\{i\})-v(S)]}
\end{equation}

The Shapley Value has found application across many fields, including law~\cite{crettez2019}, economics~\cite{moulin1992}, political science~\cite{hart1989}, and machine learning~\cite{ghorbani2020, messalas2019}. 
In this paper, we use the Shapley Value to estimate player {\em disparity} in a group during the execution of the bandit strategy.


\subsection{Multi-Armed Bandits}\label{sec:mabs}

A multi-armed bandit (MAB) problem is a type of decision problem~\cite{robbins1952some, thompson1933likelihood} in which an agent must repeatedly select from a set of mutually exclusive opportunities (called {\em arms}). The rewards that would be returned from these arms are not known to the agent at the time of selection; however, over time, through repeated selections (or {\em pulls}), the agent can build an understanding of the reward distributions available across the arms. Over its complete sequence of selections, the agent aims to maximize its total reward; therefore, it must continuously balance spending its limited pulls toward {\em exploring} the different arms to gather information about which arms are most promising versus {\em exploiting} the arm it currently believes to be the best~\cite{lattimore2018}.

Popular strategies include those of the upper confidence bound family (e.g., UCB1~\cite{auer2002}) or the $\varepsilon$-family of strategies (e.g., $\varepsilon$-greedy~\cite{lattimore2018}). In this paper, we formulate a player modeling problem as an MAB problem as in our previous work~\cite{gray2020fdg, gray2021cog}, where different game adaptation options appealing to specific player types are made available as arms to an MAB-based AI agent. As the agent selects among the adaptation options, it observes response metrics from the players (such as their self-reported motivation or daily steps) to determine which adaptations are most effective at driving those metrics. 
%

\subsection{Shapley Bandits}

In contrast to a typical bandit strategy, which only aims to maximize overall utility by selecting the arm predicted to perform the best, a Shapley Bandit tracks the contribution attributed to each player and prioritizes catering to players' preferences based on those relative contributions.



{\bf Shapley Disparity:} Given a set of players $N$, a Shapley Bandit estimates the {\em Shapley Disparity} of each player $i$ in the following way. At each step $t$, when a reward is received for an arm pull (i.e., the number of steps of a player is observed) the Shapley Value (Equation~\ref{eq:shapley}) for player $i$ is computed and is added to a {\em Cumulative Shapley Value}, $\mathit{CSV}_i$. We then define the {\em CSV Ratio}, $\mathit{CSVR}_i$, as the Cumulative Shapley Value of player $i$ divided by the sum of that of all players. In parallel, the bandit maintains a {\em Treatment Counter}, $\mathit{TC}_i$, that counts the number of times up to time $t$ that the bandit has selected the arm currently considered to be the best for player $i$. Analogously, we define the {\em TC Ratio}, $\mathit{TCR}_i$, as the sum of the Treatment Counter for player $i$ divided by the sum of that of all players. We then define the {\em Shapley Disparity} of a player $i$ at a given time $t$ as $\mathit{SD}_i = |\mathit{CSVR}_i - \mathit{TCR}_i|$. Intuitively, this measures the difference between the fraction of the contribution of a player to the group and the proportion of times the bandit has catered to that player.

Also, notice that in order to evaluate Equation~\ref{eq:shapley}, we need access to the performance of different sub-coalitions. Since not all of them are observable, in this study, we approximated them by the sum of the individual contributions (steps) of each member of the sub-coalition.

{\bf Shapley Bandit:} Given a group of players $N$, at each iteration $t$, instead of aiming to select the arm that maximizes the expected sum of step counts (as a Greedy Bandit would do), a Shapley Bandit works as follows:
\begin{itemize}
    \item {\bf Exploration:} With probability $\epsilon$ ($\epsilon = 0.01$ in our experiments\footnote{Notice this is a relatively very low exploration rate, but we set it this way because we have a forced-exploration period at the beginning of the study where during the first 9 days the bandit always explores.}), the bandit would decide to {\em explore} and choose a random arm. 
    \item {\bf Exploitation:} Otherwise, the bandit decides to {\em exploit}. The bandit then identifies which player $i$ would result in the lowest overall sum of Shapley disparities (sum of Shapley disparities of all players in $N$) if the bandit is to cater to $i$ in this turn. Then, the arm currently estimated to be the best for player $i$ will be chosen. To predict which is the best arm, the bandit uses all the observed rewards from that player up to time $t$, to estimate the expected reward of each arm. 
\end{itemize}


For example, if after 9 rounds Player 1 has $\mathit{CSV}_1 = 27,500$ steps with $\mathit{TC}_1 = 5$ and Player 2 has $\mathit{CSV}_2 = 32,800$ steps and $\mathit{TC}_2 = 4$, a Shapley Bandit would calculate a CSV ratio $\mathit{CSVR}_1 = \rfrac{27500}{(27500+32800)} = 0.456$, and TC ratio $\mathit{TCR}_1 = \rfrac{5}{(4+5)} = 0.556$ for Player 1, resulting in a Shapley Disparity of $\mathit{SD}_1 = |0.456 - 0.556| = 0.1$. If we calculate the same for Player 2, we would also obtain $\mathit{SD}_2 = 0.1$ (with a current Shapley Disparity sum of $0.1 + 0.1 = 0.2$). If we were to cater to Player 1 (and $\mathit{TC}_1$ would become = 6), the sum of their disparities (doing similar calculations) would become: $0.288$, and if we were to cater to Player 2 (and $\mathit{TC}_1$ would become = 5), the sum would become: $0.088$. A Shapley Bandit predicts Arm A to be the best for Player 1 and Arm C to be the best for Player 2. Hence, if in the next iteration, if the bandit decides to {\em exploit}, it would choose to cater to Player 2 because it has a larger Shapley Disparity, and the MAB would select Arm C for the current round (and increment Player 2's TC value).

In this way, a Shapley Bandit prioritizes the treatments allocated to players in a manner commensurate with each player's level of contribution to the team. In our case, we targeted steps to represent this contribution because in {\em Step Heroes} a player's steps serve both as a proxy for the player's real-world effort and the increased in-game utility they provide to their overall team's capability in conquering the game's challenges.

Comparing Shapley Bandits with traditional stochastic bandits, we would like to make a few observations:
\begin{itemize}
\item A Shapley Bandit directly minimizes "Shapley disparity" rather than maximizing utility. There is still an element of utility maximization once the player with the highest disparity is selected and a Shapley Bandit selects the best arm for that player; however, this is still subject to the main disparity minimization objective. Notice that minimizing Shapley disparity might sometimes go against maximizing utility (e.g., selecting the best choice for a player with a high Shapley disparity might imply choosing an arm that is not the one that would result in the highest utility gain overall). For this reason, we also note that traditional performance metrics for stochastic bandits, such as {\em regret}~\cite{auer2002}, are not directly applicable and would need redefinition.
\item A Shapley Bandit still needs to estimate which arms maximize utility for a given player in order to cater to them. Hence, it benefits from exploration in the same way as a traditional bandit.
\item A Shapley Bandit needs to maintain a series of additional quantities that traditional stochastic bandits do not need, namely a Cumulative Shapley Value and Treatment Counter for each human player. However, given the small groups of users in our study (2 human players plus one AI player), the extra computational cost is negligible.
\end{itemize}

Therefore, we do not expect a Shapley Bandit to outperform a traditional, Greedy Bandit in terms of overall maximization of the target metric (in this case, daily steps). Instead, we anticipate incurring a cost to that total utility in favor of a strategy that encourages overall player retention. 
However, we anticipate that this fairness constraint will provide a better experience to a greater number of users (measured via adherence metrics) than focusing on purely maximizing total step counts.

\section{Methods}\label{sec:methods}
We conduct a user study in the context of \textit{Step Heroes} to evaluate the effectiveness of using Shapley Bandits to personalize the social comparison environment for a small group of users in the context of a social exergame. 
Specifically, we aim to evaluate the following hypotheses:

\begin{itemize}
\item {\bf H1:} The Greedy Bandit will outperform the Shapley Bandit in increasing players’ average steps and self-reported motivation, the two metrics that the Greedy Bandit aims to maximize directly.
	\item {\bf H2:} Our Shapley Bandit will outperform the Greedy Bandit in increasing player retention and adherence with the intervention.
\end{itemize}

We recruited 46 participants (36 women, 10 men) from a major university in the U.S. For our study, healthy adults were eligible if they reported PA as being somewhat or very important to them and had access to a Fitbit health tracker or a Fitbit-compatible smartphone. We did not include any applicants that reported medical conditions or advisement that would limit their PA. We compensated participants with either gift cards or equivalent extra course credit at the end of the study. The study procedure was reviewed and approved by the Institutional Review Board (IRB) at Drexel University.

\subsection{Procedure}\label{sec:procedure}


After providing their demographic information, participants were given a set of psychology surveys, including the Iowa Netherlands Comparison Orientation Measure (INCOM-23)~\cite{gibbons1999individual} and the Identification/Contrast Scale (ICS)~\cite{buunk1997}. 
These instruments provided baselines regarding a participant's social comparison orientation. 
%
%
Participants were then provided instructions on accessing {\em Step Heroes} and linking their Fitbit accounts. Players were asked to wear their Fitbit device every day and log into the game at least once per day to synchronize their steps and participate in their daily sessions.  The first three days of the study were considered a {\em baseline period}, where we collected players' daily steps before introducing our game.


At the end of this baseline collection period, participants were ranked according to their baseline results and separated into pairs of similarly-ranked players for the rest of the intervention. Each pair was joined by an artificial player (whose steps were controlled by the game AI) to play {\em Step Heroes} as a team. Each participant could only interact with the other human player and the AI player in their team. Each team was randomly assigned one of the following three conditions: 
\begin{enumerate}
    \item {\em Control Condition}: The participants play a version of {\em Step Heroes} where the AI makes a random decision about the artificial player's steps and gameplay information each day (n=14).
    \item {\em Experimental Condition \#1}: The participants play a version of {\em Step Heroes} where the AI makes decisions based on the same Greedy Bandit strategy used in the preliminary design study (n=14).
    \item {\em Experimental Condition \#2}: The participants play a version of {\em Step Heroes} where the AI makes decisions based on a Shapley Bandit strategy (n=18).
\end{enumerate}


We asked participants to play at least once per day for 21 days.
%
%
%
We also asked them to fill out three weekly qualitative surveys to assess any technical issues and gain insight into player motivation for physical activity and their experience with the game.  
Finally, we administered an exit survey at the end of participation.


\begin{table*}[t]
	\renewcommand{\arraystretch}{1.3}
	\begin{center}
		\caption{Average Steps vs.\ Baseline for players in the Control vs.\ Greedy and Shapley Bandit conditions}
		\begin{tabular}{|c|c|c|c|c|}
			\hline
			Condition    & Steps vs.\ Baseline & P-Value vs.\ (C) & Post-Motivation & P-Value vs.\ (C) \\ \hline
			Control (C)  & -69                &                 & 1.316           &                 \\ \hline
			Greedy (E1)  & 171                & 0.1868          & 1.400           & 0.095           \\ \hline
			Shapley (E2) & -313               & 0.1023          & {\bf 1.451}     & {\bf 0.011}     \\ \hline
			\end{tabular}
		\label{tab:steps-results}
	\end{center}
\end{table*}

\subsection{Data Analysis}\label{sec:data-analysis}

For quantitative data, we use inferential statistics for analysis. Specifically, we analyzed several primary data sources resulting from the user study:

\begin{enumerate}
	\item Data related to the artificial player, including MAB predictions and selections. 
	\item User metrics regarding their interaction with the platform, including daily steps, reported motivation in the pre- and post-session questionnaire, and user participation. 
	\item Open-response feedback surveys provided to each participant following their 7\textsuperscript{th}, 14\textsuperscript{th}, and 21\textsuperscript{st} session, along with an exit survey provided to each participant on their completion of the study.
\end{enumerate}

Qualitative feedback in the weekly and exit surveys was coded by two independent researchers who used thematic analysis~\cite{braun2012thematic} to identify themes in the dataset. After initial observations, two researchers conducted a preliminary open coding. During this analysis step, both researchers reviewed each response and noted initial labels separately (e.g., the user noted a positive impact, or the user reported an increased step awareness). Then, they discussed their initial codes to form a codebook. Both researchers then independently coded the full dataset. Finally, they collaboratively discussed disagreements and modified their codes until they reached full consensus~\cite{hill2012consensual}.

\section{Results}\label{sec:results}

\subsection{Quantitative Results}

We first consider H1, where we expect that the Greedy Bandit will outperform the Shapley Bandit in the two metrics that the Greedy Bandit aims to maximize directly (i.e., average player steps and self-reported motivation). The Greedy Bandit operates without considering the fairness constraints integrated into a Shapley Bandit, and therefore we expect to observe a cost for these constraints.

Table~\ref{tab:steps-results} reports in the two left columns results on average steps for players in all three conditions relative to their baseline steps collected in the three days prior to starting their {\em Step Heroes} game. We see that the players in the Shapley Bandit condition achieved fewer steps than the Greedy Bandit condition (which was optimizing for that step count) and the random control condition. However, no condition was found to be statistically significantly different (i.e., via two-tailed T-test) from any other.


The two right columns of Table~\ref{tab:steps-results} present the average self-reported motivation following each session in all three conditions. Although the Greedy Bandit aims to maximize this metric along with steps, we find that the players in the Shapley Bandit condition reported higher post-test motivation scores. Though there was no statistically significant difference between the two experimental conditions when compared directly, the Shapley Bandit condition demonstrates a statistically significant difference between itself and the random control condition ($p{<}0.05$), whereas the Greedy Bandit did not.


We next consider H2, where we anticipate that the Shapley Bandit condition will mitigate the symptoms of the Greedy Bandit Problem. We believed this might result in lower missed sessions and lower variance in disparity scores among players in the Shapley Bandit condition. Though the data showed both of these expectations to be true---miss rates were 26\% vs.\ 25\% and average disparity scores were 0.070 vs.\ 0.067 in the Greedy Bandit and Shapley Bandit conditions, respectively---neither of these findings were statistically significant due to sample size.

\begin{figure*}[t!]
	\includegraphics[width= \textwidth]{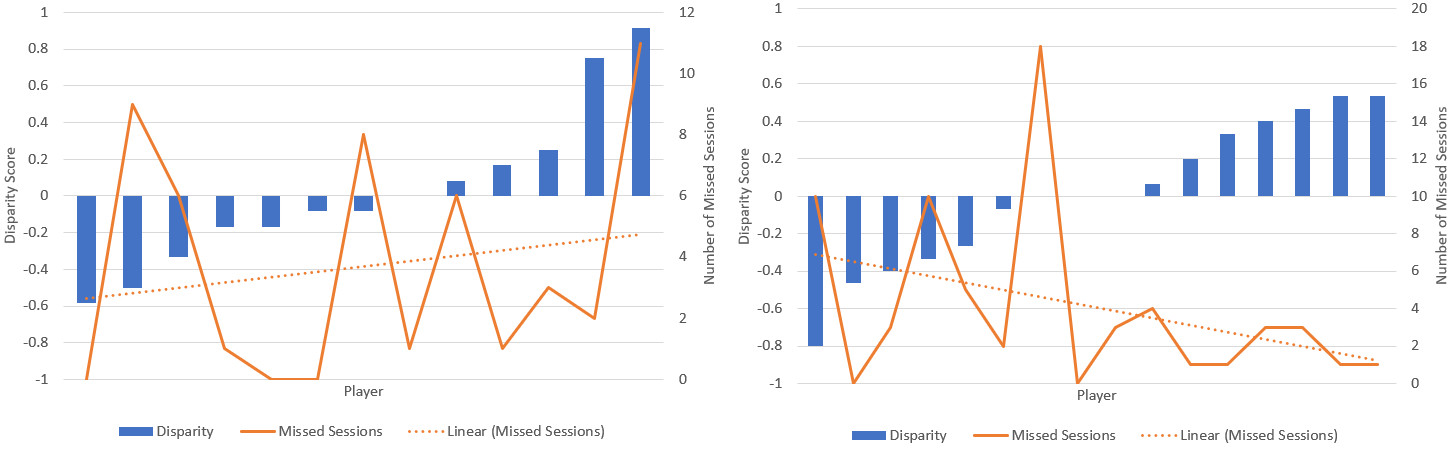}
	\centering
	\caption{Comparison of Greedy Bandit (left) to Shapley Bandit (right) disparity analysis. Compare left to Figure~\ref{fig:shapley-analysis-disparity}, reflecting the same analysis from the Greedy Bandit's deployment in the preliminary design study. As in its performance in the preliminary design study, the Greedy Bandit (left) again yields a correlation of $R{=}0.16$ between player disparity and non-adherence. In contrast, the Shapley Bandit (right) demonstrates a negative correlation coefficient of $R{=}-0.39$ that differs significantly ($p{<}0.05$)~\cite{paternoster1998} from the Greedy Bandit.}
	\label{fig:misseddaysresults}
\end{figure*}

However, we anticipated that H2 would be supported most strongly in the results of the disparity analysis that first illustrated the symptoms of the Greedy Bandit Problem. Indeed, Figure~\ref{fig:misseddaysresults} conducts the same disparity analysis on the user study data we conducted on the preliminary design study data (Section \ref{sec:greedy-bandit-problem}), with the left graph illustrating results for the Greedy Bandit and the right showing those of the Shapley Bandit. Regarding the correlation between disparity and the player's propensity to miss sessions, the Greedy Bandit performed exactly as it did in the preliminary design study, again demonstrating a Pearson's R correlation of $R{=}0.16$. In contrast, the Shapley Bandit demonstrated a negative correlation of $R{=}-0.39$. The difference between these coefficients~\cite{paternoster1998} was found to be statistically significant ($p{<}0.05$). In other words, the primary symptom of the Greedy Bandit Problem (i.e., higher disparity correlating with higher non-adherence) was confirmed in the Greedy Bandit condition of our user study scenario. However, in the Shapley Bandit condition, players experience the opposite relationship between disparity and non-adherence. 
Let us now look at our qualitative analysis, which provides further support for H2.

\subsection{Qualitative Results}\label{sec:qualitativeresults}

The research team analyzed the participants' qualitative feedback using thematic analysis~\cite{braun2012thematic}. Two coders conducted a preliminary open coding to identify notable patterns and used the consensual qualitative approach~\cite{hill2012consensual} until they reached full agreement. Consensus coding is generally more suited for small samples and for considering multiple viewpoints~\cite{mcdonald2019reliability} than inter-rater reliability (IRR).

Qualitative feedback was coded into 11 distinct themes related to participants' reasons for how they believed their participation impacted \textit{motivation for PA}, \textit{level of PA}, and \textit{group happiness}. Three additional codes were derived to capture whether or not the participant reported a positive or no effect from their response to questions asking participants to describe how participation impacted their overall \textit{motivation for PA} and \textit{level of PA}. In addition, feedback was also coded into three themes relating to other ways (besides increasing PA) participation affected their daily activities (i.e., Awareness, Relief, Step Habit). Finally, five themes emerged for how participants chose their daily profile selection. An overview of all codes is presented in Table~\ref{tab:qualitative_codebook}. 

Overall, the qualitative feedback suggested that participants believed the team they were assigned to was comprised of other participants in the study. 34 of 46 participants provided the post-study survey responses, resulting in 150 responses from the five survey questions. The weekly survey was administered three times throughout the study (i.e., Day 7, Day 14, and Day 21). 32 of 46 participants provided responses, resulting in 332 responses from the four survey questions. We consolidated the weekly and post-study survey results for questions, asking participants to rate and describe how participation impacted their  ``motivation for PA''  and ``level of PA,'' as these questions were the same for both surveys.   

\textbf{Motivation for PA:} When asked to describe how participation impacted their overall motivation for physical activity, 42\% of participants in the Shapley Bandit condition reported a positive impact versus 27\% in the Greedy Bandit condition. A majority of participants explicitly acknowledged that participation impacted their motivation to walk more or increase their step count. For example, one participant wrote, ``Yes. It makes me want to walk more so that I have more steps on my profile.'' In contrast, 69\% of participants in the Greedy Bandit condition reported a negative impact versus 42\% of participants in the Shapley Bandit condition. 

\begin{table*}[tb]
	\caption{The list of all codes and associated examples related to participants' qualitative feedback.}
    \includegraphics[width=\textwidth]{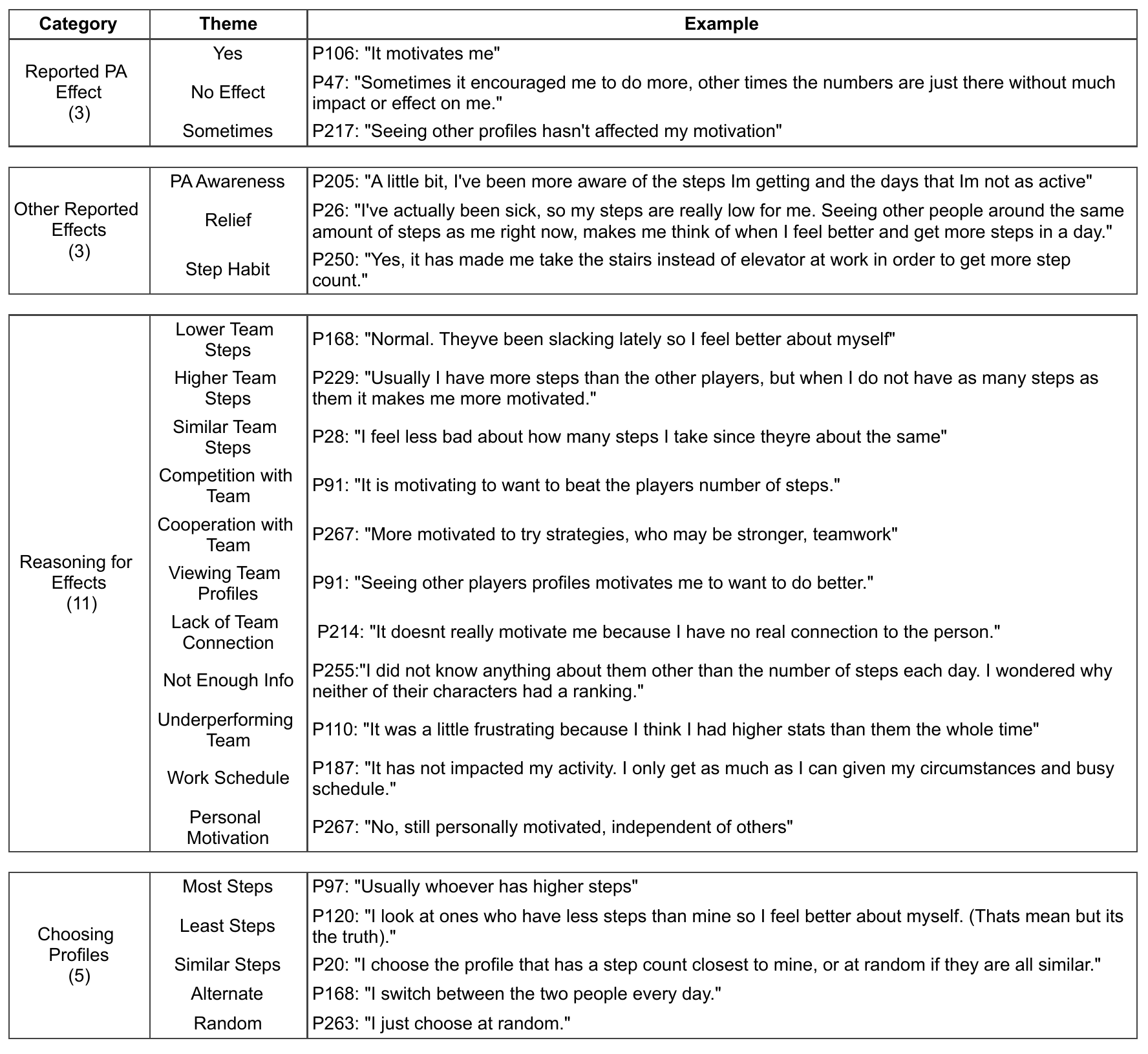}
	\centering
	\Description{}
	\label{tab:qualitative_codebook}
\end{table*}


Participants elaborated on other ways participation affected their daily activities, such as increased PA awareness (13\%) and the feeling of relief (3\%). For example, one participant wrote, ``I've actually been sick, so my steps are really low for me. Seeing other people around the same amount of steps as me right now, makes me think of when I feel better and get more steps in a day.'' 

\textbf{Level of PA:} When asked to describe how participation impacted their level of physical activity, 26\% of participants in the Shapley Bandit condition reported a positive impact in the level of PA versus 16\% in the Greedy Bandit condition. For example, one participant explained, ``Yes, prior to the study I would talk about it but not really do it, after, I actually got out there and did it, even if it was just a little.'' In contrast, 76\% of participants in the Greedy Bandit condition reported no impact versus 65\% in the Shapley Bandit condition. One participant reported, ``Not really, most of my steps are gotten between classes.''

Participants elaborated on other ways participation affected their level of PA.
For example, one participant wrote, ``\ldots I'm more aware of the steps I'm getting and the days that I'm not as active.'' 13\% described changes in their daily step habits. For instance, a participant noted, ``There were times when I chose to walk instead of finding a ride home because of the game.''
 
\textbf{Group Happiness:}  Participants were asked questions regarding their experience with their team. Using a 5-point Likert scale, they rated how happy they were with the group they were assigned within the study (0 = very unhappy, 4 = very happy). Participants in the Shapley Value condition responded with an average Likert score of 2.4, while the Greedy Bandit condition scored 2.2.

When participants were prompted to elaborate on their answers, 63\% provided a reason. For example, one participant simply wrote, ``They were fine,'' which we coded as not providing a reason. However, those who provided a reason reported step counts (42\%) as a significant factor in their team happiness. For instance, a participant wrote, ``I think that the players I was matched with took the same amount of steps so it wasn't intimidating or demotivating.'' In contrast, participants reported a lack of team connection (18\%) and insufficient information (18\%) as reasons for not being as happy with their team. For example, a participant who rated their team as ``Neutral'' wrote, ``Did not have enough info on each player to really connect with them.''

\textbf{Choosing Profiles:} When participants were asked to describe how they chose their profile each day, participants reported steps being a significant factor in their decision, ranging from choosing based on the highest steps (44\%), similar steps (14\%), or fewer steps (10\%) to their own. Alternatively, 34\% reported they chose randomly, and 7\% alternated between teammates. 

\textbf{Motivation to Login:} Using a 5-point Likert scale, participants were also asked to rate their motivation to log in to the game everyday (0 = not motivated, 4 = highly motivated). Participants in the Shapley Value condition had a Likert score of 2.3, while the Greedy Bandit condition had a score of 2.1.

These results, in general, show support for H2, where participants in the Shapley Bandit condition reported higher motivation, higher positive impact in the level of PA, and higher group satisfaction.

\section{Discussion}\label{sec:discussion}

The increasing popularity of using technology, especially AI, to personalize healthcare and to allocate resources among diverse social groups~\cite{panesar2019machine,sockolow2017risk} makes fairness a pressing issue. We believe that fairness-aware Shapley Bandit helped to mitigate the symptoms of the Greedy Bandit Problem observed in both our pre-test and our {\em Step Heroes} user study under the Greedy Bandit condition. Our hypotheses were partially supported by data resulting from our user study. We found that the Greedy Bandit outperformed the Shapley Bandit in terms of average step achievement by players, in line with our intuition that the fairness constraints would incur a penalty compared to a strategy that aimed to maximize this value at all costs (although this difference was not found to be statistically significant between the two conditions).

Our results showed that the Shapley Bandit outperformed the random control condition in terms of self-reported motivation to a statistically significant degree ($p{<}0.05$), where the Greedy Bandit was not able to achieve the same. In retrospect, this supports intuition regarding the aim of the Shapley Bandit's design toward improving player affect toward the intervention (e.g., adherence and gameplay participation) as opposed to the Greedy Bandit's aim toward maximizing specific metrics (i.e., daily steps). It may therefore be reasonable that player self-reported motivation is more closely related to the intended affect change than a behavior metric such as steps. Although this outcome was initially unexpected, it is a positive reflection on the Shapley Bandit's performance.

Our second hypothesis was fully supported by the results of our {\em Step Heroes} user study, specifically in the repeating of the disparity analysis we had conducted on our pre-test results (where the symptoms of the Greedy Bandit Problem are observed). Our disparity analysis in the {\em Step Heroes} user study data showed that 1) the Greedy Bandit performed consistently and with the same resulting correlation (Pearson's $R{=}0.16$) relating a player's disparity measurement to missed sessions even in the new context of the {\em Step Heroes} exergame, and 2) that the Shapley Bandit reduced this correlation (Pearson's $R{=}-0.39$) to a statistically significant degree ($p{<}0.05$)~\cite{paternoster1998}.

\subsection{Implications}

Our research has several implications. Concerning the literature of multi-armed bandits, the idea of introducing fairness constraints into MABs (in our case via the use of Shapley Values) in the context of group preference modeling and adaptation opens up many new avenues for future work. For example, further research is needed to study the theoretical properties of this new type of bandit and devise new measures to evaluate them beyond the traditional {\em regret} metrics, which do not fully capture fairness considerations. Notice that our idea of Shapley Bandits is not specific to exergames but is a general approach that is applicable to any domain that involves groups or teams where the bandit makes choices that affect more than one individual. Consider, for example, the idea of music recommendation for groups and task allocations in a team, as discussed in the related work.

Additionally, we show that holistic and ethical considerations of the AI's utility function are a key design issue for personalized adaptive applications. Our results provide evidence that, even though introducing fairness considerations in adaptive group exergames does not directly lead to higher PA levels, it can contribute to user retention and adherence. Adherence and long-term engagement are the foundation of whether a behavioral change intervention can have a lasting positive impact. Our research shows that if the AI only focuses on short-term performance (e.g., a group's aggregated daily steps), it risks consistently ignoring the needs of marginalized users and, over time, losing their participation. This insight has implications for the future design of player-AI interaction~\cite{zhu2021player} not only in adaptive group exergames but behavioral change apps in general.

\section{Conclusions}\label{sec:conclusion}

In this paper, we address the problem of fairness in AI-based group personalization via multi-armed bandits. Specifically, we presented evidence of what we call the {\em Greedy Bandit Problem}, arising when deploying a bandit-based AI system to personalize the group experience in a social exergame, resulting in the system focusing primarily on high-performance players and causing adherence problems. 
We use these findings to construct a new fairness-aware bandit strategy approach based on minimizing disparity estimated via the Shapley Value, which aims to align the degree of AI attention a player receives to the player's measured investment toward helping their team. We deployed that strategy in a new user study based around a team-based, idle exergame and evaluated results against two hypotheses that assessed our Shapley Bandit's performance against the traditional greedy approach.  Our results support our intuition regarding the Shapley Bandit's ability to mitigate the effects of the Greedy Bandit Problem.

We identify a few limitations of our approach, including the small sample size of this initial {\em Step Heroes} user study. Future studies with larger sample sizes may help to further establish some of the trends so far observed. Additionally, our participants are healthy, college students with relatively homogeneous health needs. Future research is needed to test our approach with a more diverse group of users. We also note that our application of the Shapley Value does not currently include all terms we wish to consider regarding user participation. The Shapley Value, and by extension the axioms that it implements, intends to capture the totality of value that an individual may bring to a group endeavor, including some benefits that may not be easily measurable. For example, we argue that a player's mere participation in a game yields value to others simply in the fact that it enables a multiplayer experience. However, this is not currently reflected in the Cumulative Shapley Value (CSV) discussed in Section~\ref{sec:shapley-bandit}, which is based solely on step performance. Solving this problem is beyond the scope of this paper, where the Shapley Value provided guidance in our effort to mitigate the effects of the Greedy Bandit Problem. However, this remains an interesting problem for future work.

Finally, there are many alternatives for modeling fairness~\cite{fairmlbook}, where we have opted for an equality-enforcing individual notion of fairness. Equality enforcing notions of fairness can have undesired side-effects (such as leveling down) in cases where there is utility dependency between the players~\cite{RahmattalabiJL+21}, as it is in our case. We leave the study of whether or not such side-effects can happen in our framework as well as studying notions of fairness within as future work.

\begin{acks}
This work is partially supported by the National Science Foundation (NSF) under Grant Numbers IIS-1816470, IIS-1917855 and Danish Novo Nordisk Foundation under Grant
Number NNF20OC0066119. The authors would like to thank all past and current members of the projects.
\end{acks}

\bibliographystyle{ACM-Reference-Format}
\bibliography{bibliography}


\end{document}